\documentclass[sigconf,nonacm]{acmart}
\setkeys{acmart.cls}{balance=false}
\setcopyright{none}
\usepackage{placeins}
\usepackage{stfloats}
\usepackage{booktabs}
\usepackage{colortbl} 
\usepackage{graphicx}
\usepackage{algorithm}
\usepackage{algpseudocode}

\definecolor{psyjevshade}{HTML}{E6F1FA}
\newcommand{\benchmark}{PsyJudgeBench}
\newcommand{\model}{PsyJev}
\newcommand{\method}{\textsc{JA-RLCD}}

\newcommand{\KL}{\operatorname{KL}}

\title{PsyJev: Psychological Analysis via System-1 Judgment}

\author{Fangyuan Liu}
\affiliation{%
  \department{State Key Laboratory of Cognitive Intelligence}
  \department{School of Computer Science and Technology}
  \institution{University of Science and Technology of China}
  \city{Hefei}
  \country{China}}
\email{liufangyuan@mail.ustc.edu.cn}

\author{Sirui Zhao}
\affiliation{%
  \department{State Key Laboratory of Cognitive Intelligence}
  \department{School of Computer Science and Technology}
  \institution{University of Science and Technology of China}
  \city{Hefei}
  \country{China}}
\email{siruit@ustc.edu.cn}

\author{Xiaohao Wang}
\affiliation{%
  \department{State Key Laboratory of Cognitive Intelligence}
  \department{School of Computer Science and Technology}
  \institution{University of Science and Technology of China}
  \city{Hefei}
  \country{China}}
\email{wangxh236@mail.ustc.edu.cn}

\author{Guo Cheng}
\affiliation{%
  \department{Computer Science}
  \institution{Beijing University of Technology}
  \city{Beijing}
  \country{China}}
\email{chengguo@emails.bjut.edu.cn}

\author{Peng Zhang}
\affiliation{%
  \department{School of Information Science and Technology}
  \institution{University of Science and Technology of China}
  \city{Hefei}
  \country{China}}
\email{zppeng@mail.ustc.edu.cn}

\author{Enhong Chen}
\affiliation{%
  \department{State Key Laboratory of Cognitive Intelligence}
  \department{School of Computer Science and Technology}
  \institution{University of Science and Technology of China}
  \city{Hefei}
  \country{China}}
\email{cheneh@ustc.edu.cn}

\renewcommand{\shortauthors}{Liu et al.}

\hypersetup{pdfauthor={Fangyuan Liu, Sirui Zhao, Xiaohao Wang, Guo Cheng, Peng Zhang, Enhong Chen}}

\makeatletter
\renewcommand{\@mkauthors}{%
  \global\setbox\mktitle@bx=\vbox{%
    \hsize=\textwidth
    \unvbox\mktitle@bx\par\medskip
    \centering
    {\large\mbox{%
      Fangyuan Liu\textsuperscript{1},
      Sirui Zhao\textsuperscript{1,*},
      Xiaohao Wang\textsuperscript{1},
      Guo Cheng\textsuperscript{2},
      Peng Zhang\textsuperscript{3},
      Enhong Chen\textsuperscript{1,*}}\par}
    \smallskip
    {\small
      \textsuperscript{1}State Key Laboratory of Cognitive Intelligence,
      School of Computer Science and Technology,\par
      University of Science and Technology of China\par
      \textsuperscript{2}Computer Science, Beijing Institute of Technology\par
      \textsuperscript{3}School of Information Science and Technology,
      University of Science and Technology of China\par}
    \smallskip
    {\small
      \texttt{\{liufangyuan,wangxh236,zppeng\}@mail.ustc.edu.cn}\quad
      \texttt{chengguo@emails.bjut.edu.cn}\par
      \texttt{\{siruit,cheneh\}@ustc.edu.cn}\quad
      \textsuperscript{*}Corresponding authors.\par}
    \bigskip}}
\makeatother

\begin{document}
\begin{abstract}
Self-disclosures on the Web support diverse psychological analyses,
from recognizing expressed experiences to identifying symptom-relevant
information. We study a shared, question-conditioned decision interface
for this heterogeneous problem in content analysis. We organize
\benchmark{}, comprising 13 annotated datasets and 82 questions
across six psychological areas, while preserving source-specific
criteria and masks for unobserved labels. We introduce \model{},
a System-1 model that directly produces distributions over answers
specified in each request. Joint supervised training establishes
the shared judgment function, followed by Judgment-Anchored RLCD
(\method{}), which adapts logit-space reward optimization to
heterogeneous psychological judgments through balanced supervision,
selective retention of correct SFT predictions, and question-group
reward optimization. The Qwen3.5-based 2B, 4B, and 9B models achieve
64.33, 66.21, and 67.56 average test F1, respectively. The 9B
model exceeds the strongest external comparison by 2.54 points.
Controlled 2B comparisons examine the training stages and anchored
refinement. Further analyses examine question-conditioned judgment,
inference cost, and probability quality, complemented by qualitative
comparisons across models.
\end{abstract}

\keywords{Psychological Analysis, Social Web, System-1 Models, Large Language Models, Psychological Analysis Benchmark}
\maketitle
\raggedbottom

\section{Introduction}

Social media, online forums, and Web communities continuously produce spontaneous accounts of personal experiences, emotional difficulties, relationships, and everyday stress. These self-disclosures provide a rich source of Web content for psychological analysis, supporting tasks ranging from recognizing expressed experiences to identifying symptom-relevant information and questionnaire answerability. However, analyzing the same content for different purposes requires distinct judgments. A post may mention a symptom without expressing it, or contain symptom-relevant information without providing enough detail to answer a questionnaire item. Such distinctions matter when organizing and analyzing Web content across different psychological tasks, where similar textual expressions may need to be interpreted under different decision criteria. This motivates a shared framework for addressing a challenge in psychological analysis of Web content: distinguishing question-specific judgments across heterogeneous datasets drawn from the social Web while preserving their source-specific decision criteria.

Research on web-based psychological analysis has produced specialized datasets and models for diverse targets, from symptom relevance and questionnaire answerability to stressors and interpersonal risk factors \citep{hengle2024angst,perez2023bdisen,gupta2022primate,mauriello2021sad,garg2023irf}. Recent work uses multi-task learning and instruction tuning to bring these tasks into shared frameworks. For instance, MentaLLaMA and the IMHI benchmark unify mental health analysis by generating joint predictions and text explanations \citep{yang2024mentallama}. The constituent tasks retain distinct label semantics, answer spaces, and decision criteria. A shared prediction interface therefore needs to preserve these distinctions while supporting categorical evaluation.

Building such a framework poses three main challenges.
First, \emph{heterogeneous judgment semantics}: different analytical questions may require distinct criteria and answer sets, even when applied to the same text. This calls for predictions conditioned on the question rather than the text alone.
Second, \emph{fragmented supervision}: existing datasets differ in annotation targets, label schemes, and coverage, complicating their integration into a shared judgment format while retaining source-specific criteria.
Third, \emph{uneven supervision}: differences in dataset sizes and label frequencies can favor well-represented tasks and frequent classes during joint training.
Addressing these challenges calls for a shared question-conditioned judgment framework and a training strategy that accounts for heterogeneous, imbalanced supervision.

To address these challenges, we introduce \benchmark{},
a benchmark for question-conditioned psychological analysis.
It organizes 13 existing datasets into 82 registered questions
across six psychological content areas. Rather than merging
heterogeneous tasks into a single label space, \benchmark{}
represents each judgment through an analytical question,
its legal answer set, and source-specific decision criteria
retained through documented label mappings. Missing-label
masking and coordinated dataset splits support joint training
and task-specific evaluation under partial supervision.
The resulting benchmark also enables evaluating how models
distinguish different judgments about the same text.

Building on this task formulation, we introduce \model{},
a System-1 model that directly predicts distributions over
question-defined answers. Given an input text, an analytical
question, and candidate options, \model{} produces a
categorical answer distribution through a shared readout,
without intermediate response generation or post-hoc category
extraction \citep{typesafe2026jev,gao2026openjevrlcd}.
To account for uneven supervision across sources, we further
propose Judgment-Anchored RLCD (\method{}), adapting
logit-space RLCD \citep{laya2026} to heterogeneous
psychological judgments. Its judgment anchor combines
source/question weighting and class-balanced supervision
with selective retention of correct SFT predictions.
A Brier-based question-group reward further refines
predictions through logit-space perturbations and
pathwise gradients. Across 2B, 4B, and 9B model scales, \model{} outperforms
the corresponding same-backbone answer-SFT baselines.
Controlled 2B comparisons show that the complete JA-RLCD
formulation outperforms continued cross-entropy training
and anchor-only refinement. Further analyses examine
question-conditioned discrimination on co-annotated texts,
inference latency, and probability quality.

Our main contributions are as follows:
\begin{itemize}
\item \textbf{A unified psychological judgment benchmark.}
We construct \benchmark{}, organizing 13 heterogeneous datasets
into a shared question-defined judgment framework that preserves
task-specific criteria and supports unified training and
evaluation across questions.
\item \textbf{System-1 modeling with judgment-anchored reward refinement.}
    We propose \model{} for direct probabilistic psychological judgment
    and \method{}, a joint refinement formulation that adapts logit-space
    RLCD through a class-balanced judgment anchor, selective SFT-reference
    retention, and question-group reward optimization under uneven
    supervision.

\item \textbf{Systematic empirical evaluation.}
    We evaluate \model{} across model scales and heterogeneous
    psychological tasks, with comparisons against benchmark-adapted
    baselines and matched training continuations.
    Ablations and further analyses assess the training components,
    question-conditioned discrimination, inference efficiency,
    and probability quality.
\end{itemize}
\section{Related Work}
\subsection{Psychological Analysis}
Psychological analysis of social Web content spans diverse tasks, including identifying depression and anxiety indications, interpersonal difficulties, self-perceptions, and emotional experiences \citep{hengle2024angst,garg2023irf,garg2023lost,garg2023lonxplain,rahman2024depressionemo}. Other resources address questionnaire answerability, symptom relevance, stressors, cognitive distortions, wellness dimensions, and crisis-related expressions \citep{gupta2022primate,perez2023bdisen,mauriello2021sad,garg2022cams,sharma2023reframing,shreevastava2021distortions,garg2024multiwd,byun2026cradle}. These tasks involve related psychological content but differ in their analytical targets and annotation criteria.
Existing models range from domain-adapted encoders such as MentalBERT to instruction-tuned language models for psychological prediction and explanation \citep{ji2022mentalbert,xu2024mentalllm,zhai2024mentalglm}. Flexible classification approaches, including GLiClass and GLiNER2, further support label- or schema-conditioned predictions \citep{stepanov2025gliclass,zaratiana2025gliner2}. MentaLLaMA and its IMHI benchmark extend psychological analysis toward multi-task instruction learning over social-media content \citep{yang2024mentallama}. Building on this direction, \benchmark{} organizes heterogeneous psychological tasks as question-defined judgments, while \model{} learns a shared direct-prediction function and refines it through \method{}.

\subsection{System-1 Decision Models}
System-1 decision models produce structured predictions without
autoregressive answer generation. Jev introduced natural-language
interfaces for probabilistic decisions \citep{typesafe2026jev},
while Laya developed typed prediction with reinforcement learning
for calibrated decisions (RLCD) \citep{laya2026}. OpenJev further
introduced dynamic-option modeling with candidate-conditioned
representations and option-set interaction \citep{openjev2026dynamic}.
Related advances include rationale-conditioned reward learning
and direct probabilistic video understanding
\citep{gao2026openjevrlcd,zhu2026videojev}.
Building on this line of work, \model{} supports heterogeneous
psychological judgments through joint supervision on \benchmark{}.
Our \method{} adapts logit-space RLCD with a class-balanced judgment
anchor, selective SFT reference retention, and question-group reward
optimization using pathwise gradients.

\section{\benchmark}
\label{sec:benchmark}
\begin{figure}[t]
\centering
\includegraphics[width=\columnwidth]{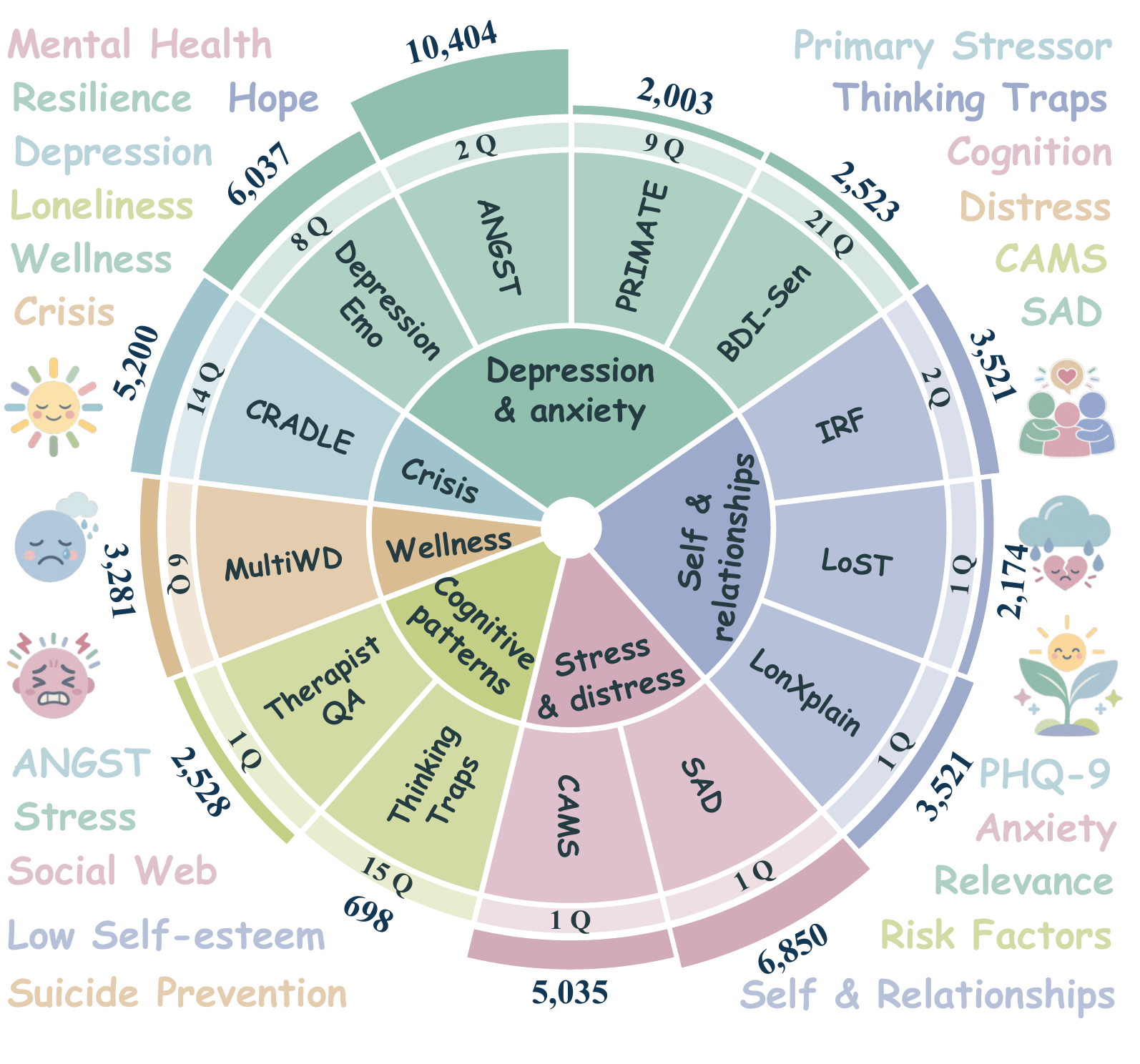}
\caption{\benchmark{} coverage. The inner groups organize 13 source datasets into six psychological content areas. Q denotes the number of questions associated with each source, and the outer bars and numbers indicate source-record counts.}
\Description{A concentric benchmark overview groups thirteen datasets into six psychological content areas and displays each dataset's question count and number of source records.}
\label{fig:coverage}
\end{figure}

\benchmark{} organizes 13 annotated datasets into 53,775 source records and 222,191 observed judgments. Figure~\ref{fig:coverage} summarizes their coverage across six thematic groups: depression/anxiety-related information, self and relationships, stress and distress, cognition, wellness, and crisis experiences. Construction involves a question registry, source-specific annotation conversion, and coordinated data partitions.

\paragraph{Question-defined tasks.}
\benchmark{} represents psychological analysis tasks as question-defined judgments, each specifying a judgment target, a decision criterion, and a set of valid answers. This formulation accommodates different analytical requirements while preserving their original annotation criteria. For example, PRIMATE assesses whether a text provides enough information to answer a PHQ-9 item, BDI-Sen evaluates sentence relevance to a BDI-II symptom \citep{gupta2022primate,perez2023bdisen}, and MultiWD identifies mentions of specific wellness dimensions. Each judgment is represented as $(x,q,\mathcal A_q,y)$, comprising the source text, question, valid answer set, and observed answer. The resulting registry contains 82 questions, including 79 binary and three multiclass questions. 

\paragraph{Annotation harmonization.}
Following these question definitions, we convert heterogeneous source annotations into observed judgments while retaining their original labeling criteria. Binary and categorical labels are mapped to their corresponding answer sets, while multi-label and category--time annotations are expanded into separate questions. For Thinking Traps, we normalize category names and aggregate repeated completions to recover 15-category label sets. For CRADLE, seven crisis categories combined with ongoing/past status yield 14 questions. For TherapistQA, we retain the dominant distortion category and use only the patient's question, excluding therapist responses and annotated spans. After conversion, we check for conflicting annotations associated with the same text and judgment target. We remove two TherapistQA records with identical questions but conflicting dominant labels and mask 22 conflicting judgments across 11 same-text/same-target keys. Unaffected judgments remain available for supervision, while source identifiers and annotation provenance, including gold/silver distinctions, are preserved.

\paragraph{Partial supervision and coordinated splits.}
The harmonized records provide supervision only for questions with observed annotations; unannotated questions are excluded rather than treated as negative examples. To construct consistent evaluation partitions, we coordinate records sharing identical texts or available source-group identifiers and group identified near-duplicates to reduce overlap across splits. When grouping conflicts with the original partitions, held-out assignments take precedence. The resulting benchmark contains 33,511 training, 5,852 validation, and 14,412 test source records. These counts refer to source records rather than unique texts. Further details on data provenance, partitioning, and duplicate checks are provided in Appendix~\ref{app:data}.

\begin{figure*}[t]
\centering
\includegraphics[width=\textwidth]{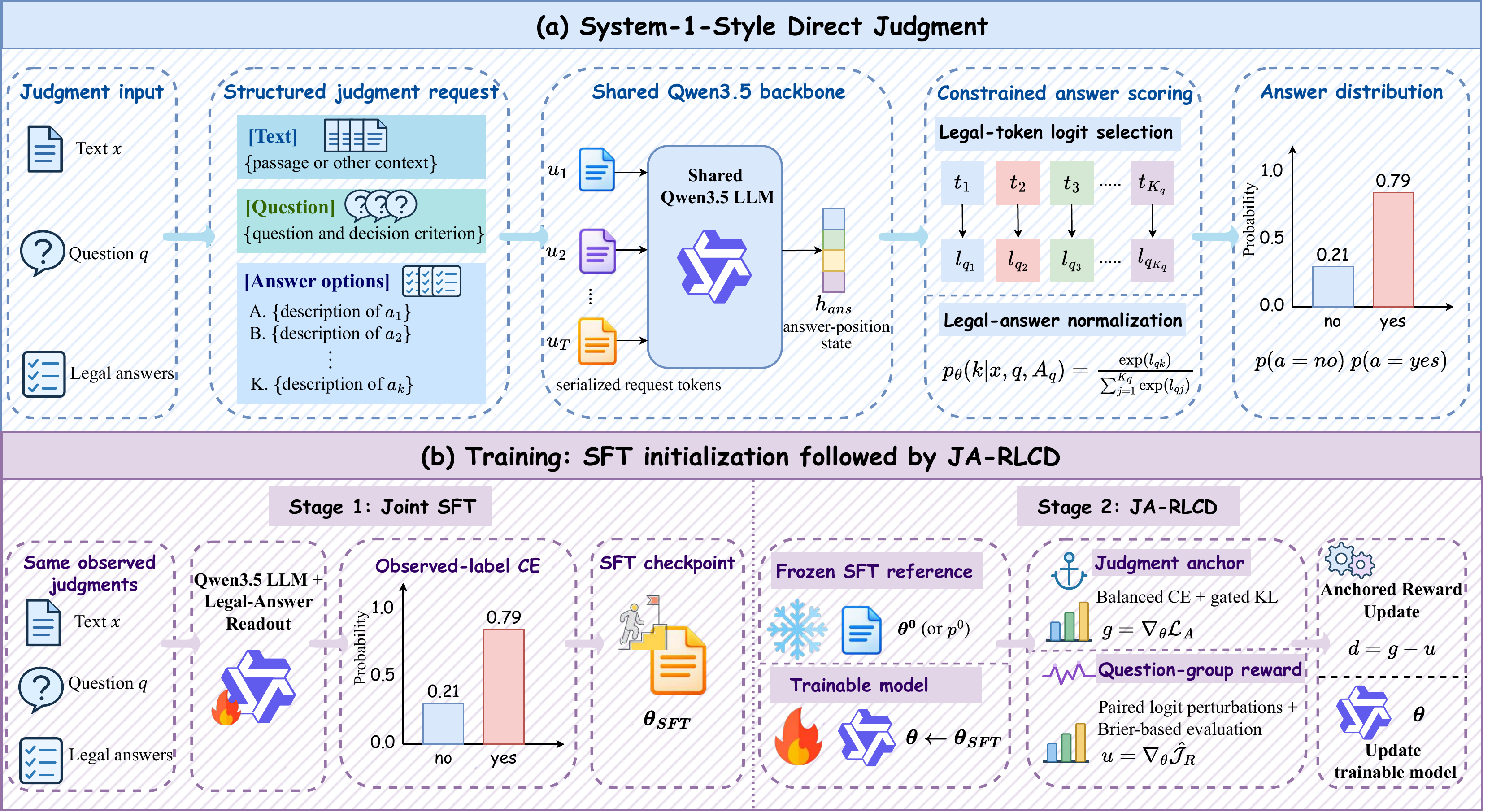}
\caption{\model{} overview. (a) System-1-style direct judgment through question-conditioned input serialization and constrained legal-answer scoring. (b) Two-stage training with joint SFT followed by JA-RLCD, combining judgment anchoring with logit-space exploration and question-group reward optimization.}
\Description{The prediction pathway serializes the text, question, criterion, and labeled candidates, encodes them with a Qwen3.5 backbone, and normalizes answer-position vocabulary logits over the legal answer identifiers. The training pathway applies supervised fine-tuning followed by JA-RLCD, which explores answer logits, evaluates Brier-based question-group rewards, and updates the judgment policy with pathwise reward gradients and a judgment anchor.}
\label{fig:method}
\end{figure*}

\section{PsyJev}
\label{sec:model}
PsyJev is a System-1-style model for unified psychological judgment. Given a text, an analytical question, and its legal answers, PsyJev directly produces a categorical distribution through a shared, question-conditioned readout, without autoregressive answer generation. Its training proceeds in two stages: joint supervised fine-tuning (SFT) learns the shared decision function across heterogeneous tasks, followed by Judgment-Anchored RLCD (\method{}), which combines balanced supervision, selective SFT-reference retention, and question-group reward optimization through logit-space exploration. Figure~\ref{fig:method} illustrates the overall framework.

\subsection{System-1-Style Direct Judgment}
\label{sec:prediction-pipeline}

PsyJev implements direct psychological judgment through a question-conditioned, constrained prediction interface. Rather than generating and parsing free-form responses, it evaluates the admissible answers using their corresponding language-model logits and directly produces a normalized probability distribution.

\paragraph{Question-conditioned representation.}
Given a text $x$, an analytical question $q$, and its legal answer set $\mathcal A_q=\{a_1,\ldots,a_{K_q}\}$, PsyJev constructs a structured input that combines the source text, question, decision criterion, and candidate descriptions. Each candidate is assigned a distinct answer identifier, allowing different psychological judgments to be expressed through the same input format. A shared Qwen3.5 backbone processes the complete judgment request and produces a contextual representation at the answer position.

\paragraph{Constrained answer scoring.}
Instead of generating an answer sequence, PsyJev directly extracts the vocabulary logits corresponding to the admissible answer identifiers. Let $t_k$ denote the token assigned to candidate $a_k$, and let $z_\theta(t_k\mid x,q,\mathcal A_q)$ denote its logit at the answer position. The resulting judgment distribution is
\begin{equation}
\begin{aligned}
\ell_{qk}
&=z_\theta(t_k\mid x,q,\mathcal A_q),\\
p_\theta(k\mid x,q,\mathcal A_q)
&=\frac{\exp(\ell_{qk})}
{\sum_{j=1}^{K_q}\exp(\ell_{qj})}.
\end{aligned}
\label{eq:readout}
\end{equation}

Normalization is restricted to the legal answer set, yielding a categorical distribution whose dimensionality is determined by the current question.

\paragraph{Direct probabilistic judgment.}
The predicted answer is obtained directly from the normalized answer distribution:
\begin{equation}
\hat y=\arg\max_{k\in\{1,\ldots,K_q\}}
p_\theta(k\mid x,q,\mathcal A_q).
\label{eq:prediction}
\end{equation}

The same scoring procedure supports binary and multiclass psychological judgments without task-specific classification heads or autoregressive answer generation. By retaining the complete distribution over admissible answers, PsyJev also provides the probabilistic decision interface used by subsequent JA-RLCD refinement. Appendix~\ref{app:implementation} provides the implementation details.

\subsection{Supervised Fine-Tuning}
\label{sec:sft}
Joint SFT establishes the shared judgment function by learning from observed annotations across source datasets. Let $\mathcal D_s$ denote the records from source $s$, and let $\mathcal O(x)$ denote the questions annotated for record $x$. We minimize categorical cross-entropy:
\begin{equation}
\mathcal L_{\rm SFT}
=
-\frac{1}{S}\sum_{s=1}^{S}
\mathbb E_{x\sim\mathcal D_s}
\left[
\frac{1}{|\mathcal O(x)|}
\sum_{q\in\mathcal O(x)}
\log p_\theta(y_q\mid x,q,\mathcal A_q)
\right].
\label{eq:sft}
\end{equation}

The outer average assigns equal weight to each source, while the inner average prevents records with more annotated questions from dominating the objective. Missing question labels contribute no loss. The resulting checkpoint $\theta_{\rm SFT}$ initializes the subsequent JA-RLCD refinement.

\subsection{Judgment-Anchored RLCD}
\label{sec:ja-rlcd}
RLCD explores probabilistic decisions through logit-space perturbations and reward evaluation \citep{laya2026}. Building on this formulation, \method{} addresses uneven source, question, and answer frequencies in psychological judgments through a joint refinement objective: a judgment anchor combines class-balanced supervision and selective retention of correct SFT predictions, while a Brier-based reward emphasizes higher-error question groups within each microbatch. Starting from the SFT checkpoint, we treat the legal-answer distribution as a judgment policy and refine it through paired Gaussian perturbations and differentiable reward optimization. For each source-specific microbatch of observed judgments, reward gradients are computed through pathwise differentiation and combined with the anchor gradients before accumulation and AdamW updates.  Algorithm~\ref{alg:ja-rlcd} summarizes the procedure.

\begin{algorithm}[t]
\caption{Judgment-Anchored RLCD (JA-RLCD)}
\label{alg:ja-rlcd}
\small
\begin{algorithmic}[1]
\Require Pretrained LLM $\theta_{\rm pre}$; annotated records $\mathcal D$
\Ensure Trained direct-judgment model $\theta$
\State $\theta_{\rm SFT}\gets\operatorname{SFT}(\theta_{\rm pre},\mathcal D)$ using Eq.~\ref{eq:sft}
\State Initialize $\theta\gets\theta_{\rm SFT}$; freeze reference $\theta_0\gets\theta_{\rm SFT}$
\State Compute class costs $c_i$ and source/question training counts
\For{each record batch $\mathcal C\subset\mathcal D$}
  \State Set $a_i$ using batch size $B=|\mathcal C|$ \Comment{Appendix~\ref{app:implementation}}
  \State $d_{\rm acc}\gets0$
  \For{each source-specific pack $\mathcal B$ of at most 16 judgments}
    \State $\ell\gets\operatorname{Readout}(\operatorname{LLM}_\theta,\mathcal B)$ \Comment{Eq.~\ref{eq:readout}}
    \State $p_0\gets\operatorname{ReadoutProb}(\operatorname{LLM}_{\theta_0},\mathcal B)$ \Comment{Frozen}
    \State Set $m_i\gets\mathbb{I}[\arg\max_k p_{0,i,k}=y_i]$ for $i\in\mathcal B$
    \State Form balanced CE and gated KL anchor $\mathcal L_{\rm A}$ \Comment{Eq.~\ref{eq:anchor}}
    \State Set $A_{\mathcal B}\gets\sum_{i\in\mathcal B}a_i$ and $\mathcal G\gets\{q_i:i\in\mathcal B\}$
    \State Explore with two Gaussian directions per judgment; form $z_1^\pm,z_2^\pm$
    \State Compute legal-answer probabilities $\mathcal P_1^\pm,\mathcal P_2^\pm$ \Comment{Eq.~\ref{eq:smoothing}}
    \State Evaluate $\mathcal R_\tau(\mathcal P_m^v)$ on $\mathcal B$ for all four draws
    \State $\widehat{\mathcal J}_{\rm R}\gets\frac{A_{\mathcal B}}{4}\sum_{m=1}^{2}\sum_{v\in\{+,-\}}\mathcal R_\tau(\mathcal P_m^v)$
    \State $g\gets\nabla_\theta\mathcal L_{\rm A}$; $u\gets\nabla_\theta\widehat{\mathcal J}_{\rm R}$ \Comment{Pathwise}
    \State $d_{\mathcal B}\gets g-u$ \Comment{Anchored refinement; Eq.~\ref{eq:gradient}}
    \State $d_{\rm acc}\gets d_{\rm acc}+d_{\mathcal B}$
  \EndFor
  \State $\theta\gets\operatorname{AdamW}(\theta,\operatorname{ClipNorm}(d_{\rm acc},1))$
\EndFor
\State \Return $\theta$ \Comment{Use direct readout for inference}
\end{algorithmic}
\end{algorithm}

\paragraph{Judgment anchor.}
\label{sec:balanced-anchoring}
For judgment $i\in\mathcal B$, training-count weight $a_i$ balances source/question observation frequency, and cost $c_i=c_{q_i,y_i}$ balances answer classes (Appendix~\ref{app:implementation}). Let $p_i$ and $p_{0,i}$ denote the trainable and frozen distributions over $\mathcal A_{q_i}$. The anchor is:
\begin{equation}
\mathcal L_{\rm A}=\sum_{i\in\mathcal B}a_i c_i[-\log p_i(y_i)]
+\lambda\sum_{i\in\mathcal B}a_i m_i\KL(p_{0,i}\|p_i),
\label{eq:anchor}
\end{equation}
where $m_i=\mathbb{I}[\arg\max_k p_{0,i,k}=y_i]$. The indicator $\mathbb{I}[\cdot]$ equals 1 when its condition holds and 0 otherwise, so reference retention applies only to judgments the frozen model answers correctly. The label term anchors reward-driven changes to observed answers, while the reference term encourages retention of the SFT distribution where its prediction agrees with the annotation.

\paragraph{Brier-based question-group reward.}
\label{sec:question-group-scoring}
Let $\mathcal G=\{q_i:i\in\mathcal B\}$ be the observed questions in a microbatch, and let $\mathcal P=(p_i)_{i\in\mathcal B}$ collect its legal-answer distributions. We define each question's reward as the negative class-weighted Brier error \citep{gneiting2007proper}, then aggregate the question rewards with a mean and an entropic soft minimum:
\begin{equation}
\begin{aligned}
R_q(\mathcal P)&=-\frac{\sum_{i\in\mathcal B:q_i=q}c_i\,\tfrac12\|p_i-e_{y_i}\|_2^2}{\sum_{i\in\mathcal B:q_i=q}c_i},\\
\mathcal R_\tau(\mathcal P)&=\tfrac12\overline R-\frac1{2\tau}\log\left(\frac1{|\mathcal G|}\sum_{q\in\mathcal G}e^{-\tau R_q(\mathcal P)}\right).
\end{aligned}
\label{eq:reward}
\end{equation}

Here $e_{y_i}$ is the one-hot answer and $\overline R=|\mathcal G|^{-1}\sum_{q\in\mathcal G}R_q(\mathcal P)$. Higher reward indicates lower prediction error. With $\tau=4$, the soft-minimum term places more weight on questions with lower reward within the current microbatch. This aggregate is the negative of the corresponding question-group risk.

\paragraph{Logit-space exploration.}
\label{sec:score-optimization}
We draw $M=2$ independent Gaussian directions $\epsilon_{m,i}\sim\mathcal N(0,I_{K_i})$ per judgment, where $K_i=|\mathcal A_{q_i}|$. The projection $P_i=I-\mathbf1\mathbf1^\top/K_i$ removes the softmax-invariant common shift; $\mathbf1\in\mathbb R^{K_i}$ denotes the all-ones vector. With $A_{\mathcal B}=\sum_{i\in\mathcal B}a_i$, paired perturbations give the sampled, Gaussian-smoothed reward:
\begin{equation}
\begin{aligned}
z_{m,i}^{\pm}&=\ell_{\theta,i}\pm\sigma P_i\epsilon_{m,i},\\
p_{m,i}^{v}&=\operatorname{softmax}(z_{m,i}^{v}),\qquad v\in\{+,-\},\\
\widehat{\mathcal J}_{\rm R}&=\frac{A_{\mathcal B}}{2M}\sum_{m=1}^{M}\sum_{v\in\{+,-\}}\mathcal R_\tau(\mathcal P_m^v).
\end{aligned}
\label{eq:smoothing}
\end{equation}

Here $\mathcal P_m^v=(p_{m,i}^v)_{i\in\mathcal B}$, and softmax is over each judgment's legal answers. All four reward evaluations recompute Eq.~\ref{eq:reward} on the same microbatch from the same model logits. The individual anchor weights $a_i$ balance judgments; their sum $A_{\mathcal B}$ scales the entire reward estimate rather than reweighting its question groups.

\paragraph{Pathwise reward gradient.}
Since the question-group reward is differentiable with respect to the answer distributions, we directly propagate its gradients through the perturbed logits and legal-answer softmax. With Gaussian noise sampled independently of $\theta$, this yields the pathwise reward gradient $u=\nabla_\theta\widehat{\mathcal J}_{\rm R}$, allowing the sampled rewards to guide refinement without relying on a score-function estimator. We further examine the choice of gradient estimator in Section~\ref{sec:risk-ablation}.

\paragraph{Anchored policy refinement.}
\label{sec:controlled-update}
We maximize reward while maintaining the judgment anchor. Writing $g=\nabla_\theta\mathcal L_{\rm A}$, the anchored objective and its descent direction are:
\begin{equation}
\begin{aligned}
\widehat{\mathcal J}_{\rm AP}&=\widehat{\mathcal J}_{\rm R}-\mathcal L_{\rm A},\\
d&=-\nabla_\theta\widehat{\mathcal J}_{\rm AP}=g-u.
\end{aligned}
\label{eq:gradient}
\end{equation}

We accumulate $d$ across microbatches, clip the accumulated direction to a global norm of 1, and apply AdamW, keeping the SFT reference frozen. Equivalently, the anchored objective minimizes a supervised loss combining the judgment anchor with Gaussian-smoothed, question-group prediction risk.

\section{Experiments}
\label{sec:experiments}

\subsection{Experimental Setup}
\label{sec:setup}
\subsubsection{Benchmark and Metrics}
\label{sec:benchmark-metrics}
We evaluate PsyJev on \benchmark{} using the coordinated splits described in Section~\ref{sec:benchmark}. We report positive-class F1 for binary questions and macro-F1 for multiclass questions, averaging across observed questions within each dataset. The overall score equally weights the 13 datasets. We assess probability quality using normalized negative log-likelihood (NLL), Brier score, and expected calibration error (ECE) with 15 equal-width bins. NLL is normalized by $\log |\mathcal A_q|$ to account for varying answer-set sizes. Efficiency is measured by model computation time per judgment.

\subsubsection{Baselines}
\label{sec:comparison-baselines}
We compare with 18 benchmark-adapted baselines and the zero-shot Jev API. The adapted baselines fall into three groups. Compact classification and decision models include MentalBERT, MentalRoBERTa, ModernBERT-large, GLiClass, GLiNER2, and Laya-multilingual \citep{ji2022mentalbert,warner2024modernbert,stepanov2025gliclass,zaratiana2025gliner2,laya2026}. Laya uses its released architecture and checkpoint; its reported result follows benchmark SFT and continuation with its native RLCD objective. The 2--4B language models are FLAN-T5-XL, Qwen3.5-2B/4B, Llama-3.2-3B, Phi-4-mini, and Gemma-3-4B \citep{chung2022flan,qwen2026qwen35,meta2024llama32,microsoft2025phi4mini,gemmateam2025gemma3}. The 7--9B models include MentaLLaMA-7B, Qwen2.5-7B, Llama-3.1-8B, Qwen3-8B, Gemma-2-9B, and Qwen3.5-9B \citep{yang2024mentallama,qwen2024,dubey2024llama,qwen2025qwen3,gemmateam2024gemma2,qwen2026qwen35}. All 18 models are adapted on the benchmark training partition. Classification and decision models predict the observed task labels directly. Language models learn to produce the corresponding answers and are evaluated by scoring the allowed answer sequences. We additionally evaluate Typesafe's Jev (\texttt{jev-1.14.0}) through its official API without benchmark fine-tuning or demonstrations \citep{typesafe2026jev}. Appendix~\ref{app:baselines} specifies both adaptation and API protocols.

\subsubsection{Implementation Details}
\label{sec:training-details}
We initialize \model{}-2B/4B/9B from the corresponding Qwen3.5 text backbones and train each model with joint SFT followed by JA-RLCD, using one epoch per stage. Source texts are truncated to 512 tokens, with batches of up to 64 texts. The 2B model updates all parameters at a learning rate of $2\times10^{-6}$ in both stages. The 4B/9B models use LoRA (rank 16, scaling 32, zero dropout), with learning rates of $2\times10^{-4}$ for SFT and $2\times10^{-6}$ for JA-RLCD. All models use AdamW with weight decay 0.01, gradient clipping at 1.0, and 3\% warmup followed by cosine decay. Backbone computation uses BF16, and the final checkpoint of each stage is retained. We measure model computation time using a single H100 80GB GPU, one text per request, and a fixed 13-text training probe shared by all timed models; classification metrics use the complete test partition. Further details appear in Appendices~\ref{app:implementation} and~\ref{app:timing}.
\begin{table*}[htbp]
\caption{Performance comparison on \benchmark{} (F1, \%).
Avg. denotes the unweighted mean across 13 datasets.
Bold indicates the best result within each benchmark-adapted
model group. Jev is evaluated zero-shot.
Latency measures model computation (ms/judgment) on the fixed H100 CUDA Graph probe.
Abbreviations: DepEmo (DepressionEmo), BDISen (BDI-Sen),
LonXpl (LonXplain), ThinkTrap (Thinking Traps),
and TherapQA (TherapistQA).}
\label{tab:main}
\centering\fontsize{8.1}{9.4}\selectfont
\setlength{\tabcolsep}{1pt}
\renewcommand{\arraystretch}{1.10}

\definecolor{resultgroup1}{HTML}{90bfad}
\definecolor{resultgroup2}{HTML}{9daacb}
\definecolor{resultgroup3}{HTML}{d2abbb}
\definecolor{resultgroup4}{HTML}{c2cf84}
\definecolor{resultgroup5}{HTML}{d9bc91}
\definecolor{resultgroup6}{HTML}{a0c4cd}

\definecolor{modelgroupbg}{HTML}{F1F3F5}
\definecolor{psyjevshade}{HTML}{E6F1FA}

\begingroup
\setlength{\tabcolsep}{3.8pt}
\renewcommand{\arraystretch}{1.12}

\resizebox{\textwidth}{!}{%
\begin{tabular}{@{}l*{15}{c}@{}}
\toprule
& \multicolumn{4}{c}{\textcolor{resultgroup1!45!black}{\textbf{Depression / Anxiety}}}
& \multicolumn{3}{c}{\textcolor{resultgroup2!45!black}{\textbf{Self / Relationships}}}
& \multicolumn{2}{c}{\textcolor{resultgroup3!45!black}{\textbf{Stress / Distress}}}
& \multicolumn{2}{c}{\textcolor{resultgroup4!45!black}{\textbf{Cognition}}}
& \textcolor{resultgroup5!45!black}{\textbf{Wellness}}
& \textcolor{resultgroup6!45!black}{\textbf{Crisis}}
& \textbf{Avg.}
& \textbf{Inference} \\
\cmidrule(lr){2-5}
\cmidrule(lr){6-8}
\cmidrule(lr){9-10}
\cmidrule(lr){11-12}
\cmidrule(lr){13-13}
\cmidrule(lr){14-14}
\cmidrule(lr){16-16}
\textbf{Model}
& \textbf{DepEmo}
& \textbf{ANGST}
& \textbf{PRIMATE}
& \textbf{BDISen}
& \textbf{IRF}
& \textbf{LoST}
& \textbf{LonXpl}
& \textbf{SAD}
& \textbf{CAMS}
& \textbf{ThinkTrap}
& \textbf{TherapQA}
& \textbf{MultiWD}
& \textbf{CRADLE}
& \textbf{Avg.}
& \textbf{Latency (ms)} \\
\midrule
\rowcolor{modelgroupbg}
\multicolumn{16}{@{}l}{\textit{Classification and Decision Models}} \\
\addlinespace[2pt]
MentalBERT & 52.09 & 43.38 & 27.03 & 8.61 & 65.00 & 17.10 & 77.43 & 66.60 & 38.46 & 0.00 & 7.82 & 37.88 & 0.00 & 33.95 & 0.46 \\
MentalRoBERTa & 60.74 & 58.21 & 33.11 & 5.13 & 75.45 & 31.55 & 80.66 & 67.87 & 39.06 & 0.00 & 10.51 & 38.69 & 0.00 & 38.54 & 0.46 \\
ModernBERT-large & 68.95 & 64.22 & 48.06 & 42.90 & 82.65 & 67.12 & 83.17 & 71.31 & 46.76 & 15.47 & 19.25 & 53.49 & 20.08 & 52.57 & 1.16 \\
GLiClass & 75.78 & 69.70 & 56.97 & \textbf{57.36} & 83.41 & 68.01 & 84.77 & 73.47 & \textbf{49.66} & 12.95 & \textbf{19.62} & 62.99 & \textbf{38.88} & 57.97 & 1.90 \\
GLiNER2 & \textbf{75.90} & \textbf{72.24} & \textbf{68.72} & 46.71 & \textbf{84.44} & \textbf{70.56} & \textbf{85.92} & \textbf{73.82} & 49.56 & \textbf{16.60} & 19.51 & \textbf{65.24} & 37.88 & \textbf{59.01} & 3.16 \\
Laya& 68.61 & 71.47 & 50.46 & 35.30 & 77.19 & 58.10 & 79.34 & 70.85 & 39.59 & 0.00 & 9.42 & 50.90 & 2.39 & 47.20 & 1.31 \\
\midrule
\rowcolor{modelgroupbg}
\multicolumn{16}{@{}l}{\textit{2--3B Models}} \\
\addlinespace[2pt]
FLAN-T5-XL & 71.30 & \textbf{72.98} & 57.52 & 41.21 & 63.30 & 66.22 & 84.56 & 74.18 & 42.27 & 9.37 & 8.81 & 56.40 & 39.29 & 52.88 & 10.33 \\
Qwen3.5-2B & 75.74 & 70.05 & \textbf{72.83} & 50.77 & \textbf{84.82} & 69.91 & 85.48 & 73.56 & 50.94 & 23.86 & 19.45 & 63.64 & 43.09 & 60.32 & 8.44 \\
Llama-3.2-3B & 77.57 & 68.88 & 65.94 & 52.18 & 84.25 & 69.49 & \textbf{86.10} & 73.55 & 50.03 & 22.64 & 19.59 & 65.34 & 48.57 & 60.32 & 18.82 \\
\rowcolor{psyjevshade}
\textbf{\model{}-2B} & \textbf{78.25} & 70.36 & 69.26 & \textbf{55.18} & 83.97 & \textbf{70.38} & 85.31 & \textbf{74.72} & \textbf{51.47} & \textbf{48.58} & \textbf{25.38} & \textbf{68.34} & \textbf{55.10} & \textbf{64.33} & 4.38 \\
\midrule
\rowcolor{modelgroupbg}
\multicolumn{16}{@{}l}{\textit{4B Models}} \\
\addlinespace[2pt]
Phi-4-mini & 74.22 & 67.77 & 63.41 & 52.67 & 83.77 & 66.36 & 85.61 & 75.01 & 50.03 & 14.24 & 15.62 & 62.48 & 43.55 & 58.06 & 18.39 \\
Gemma-3-4B-IT & 76.85 & 69.14 & 70.70 & 54.50 & 84.55 & 69.98 & 86.26 & 74.84 & 50.85 & 22.98 & 16.35 & 66.26 & 46.62 & 60.76 & 23.82 \\
Qwen3.5-4B & 77.27 & 67.79 & \textbf{72.50} & \textbf{60.74} & 85.44 & 68.92 & \textbf{86.98} & 75.80 & 53.36 & 33.14 & 27.93 & 67.22 & 61.40 & 64.50 & 25.25 \\
\rowcolor{psyjevshade}
\textbf{\model{}-4B} & \textbf{78.63} & \textbf{69.67} & 71.76 & 56.23 & \textbf{85.54} & \textbf{71.37} & 86.59 & \textbf{76.10} & \textbf{53.88} & \textbf{50.17} & \textbf{29.80} & \textbf{69.54} & \textbf{61.43} & \textbf{66.21} & 12.01 \\
\midrule
\rowcolor{modelgroupbg}
\multicolumn{16}{@{}l}{\textit{7--9B Models}} \\
\addlinespace[2pt]
Qwen2.5-7B & 75.77 & 67.10 & 69.76 & 58.17 & 85.25 & 70.77 & 86.57 & 75.71 & 50.90 & 33.41 & 27.37 & 65.58 & 50.06 & 62.80 & 31.52 \\
Llama-3.1-8B & 76.72 & 68.31 & 70.87 & 59.21 & \textbf{85.52} & 70.64 & \textbf{87.02} & 77.01 & 49.77 & 30.37 & 28.10 & 65.39 & 58.67 & 63.66 & 35.18 \\
Qwen3-8B & 75.64 & 68.84 & 68.45 & 55.92 & 84.38 & 70.98 & 86.92 & 74.84 & \textbf{54.54} & 27.77 & 26.23 & 66.16 & 47.51 & 62.17 & 35.94 \\
Gemma-2-9B-IT & 77.34 & 69.10 & 71.34 & 59.75 & 85.40 & 70.43 & 86.67 & 76.45 & 52.16 & 39.83 & 23.14 & 66.49 & \textbf{67.17} & 65.02 & 45.82 \\
Qwen3.5-9B & 75.19 & 69.81 & 68.95 & 54.50 & 85.49 & 71.24 & 86.29 & 77.27 & 51.88 & 37.21 & 24.37 & 64.66 & 60.97 & 63.68 & 36.27 \\
MentaLLaMA-7B & 75.27 & \textbf{69.88} & 63.02 & 54.84 & 84.44 & 70.28 & 85.89 & 75.83 & 48.93 & 2.95 & 17.45 & 63.94 & 46.43 & 58.40 & 34.12 \\
\rowcolor{psyjevshade}
\textbf{\model{}-9B} & \textbf{78.03} & 69.05 & \textbf{71.75} & \textbf{63.87} & 85.01 & \textbf{71.52} & 86.41 & \textbf{77.81} & 54.52 & \textbf{54.37} & \textbf{33.22} & \textbf{69.37} & 63.36 & \textbf{67.56} & 17.12 \\
\midrule
\rowcolor{modelgroupbg}
\multicolumn{16}{@{}l}{\textit{Closed API Model (Zero-Shot)}} \\
\addlinespace[2pt]
Jev (API) & 66.91 & 60.19 & 50.97 & 48.42 & 62.25 & 46.23 & 81.84 & 70.48 & 38.26 & 35.04 & 21.62 & 57.57 & 48.07 & 52.91 & \textcolor{black!32}{\textemdash} \\
\bottomrule
\end{tabular}%
}
\endgroup

\end{table*}

\subsection{Performance Evaluation}
\label{sec:performance}
Table~\ref{tab:main} shows that \model{} consistently achieves the highest average F1 within each model-size group, reaching 64.33, 66.21, and 67.56 at 2B, 4B, and 9B, respectively. Compared with the corresponding Qwen3.5 answer models, these results represent gains of 4.01, 1.71, and 3.88 points. Notably, the 4B and 9B variants also outperform the strongest external baseline, despite the smaller size of the former. Across datasets, PsyJev shows notable gains on tasks involving
cognitive distortions and wellness-related judgments,
although gains vary across question criteria and are not
uniform across all datasets. The general-purpose Jev API achieves 52.91 average F1 in zero-shot evaluation, while the benchmark-adapted \model{} variants perform substantially better, providing a zero-shot reference alongside the benchmark-adapted comparisons. Overall, these results demonstrate the effectiveness of \model{} across heterogeneous psychological judgment tasks and model scales.

\subsection{Ablation Study}
\label{sec:ablation}

Table~\ref{tab:controls} uses the 2B model to examine training stages, JA-RLCD components, and matched-budget prediction formats. Direct-scoring continuations share the same SFT initialization and schedule; the answer-sequence control uses its corresponding SFT model with the same update budget.

\subsubsection{Training Stages}
\label{sec:refinement-ablation}

JA-RLCD achieves 64.33 average F1, improving upon SFT by 3.65 points. It also outperforms continued cross-entropy (CE) training and adapted RLCD by 2.61 and 5.46 points, respectively. Continued CE controls for additional training, while adapted RLCD combines CE supervision with the original RLCD's noisy-logit score-function objective. These comparisons show that additional training alone does not explain the improvement and demonstrate the effectiveness of the overall JA-RLCD formulation. Since adapted RLCD differs in both its reward formulation and gradient estimator, the comparison does not isolate the contribution of either component.

\begin{table}[t]
\caption{2B ablations and matched-budget controls. $\Delta$F1 is relative to full JA-RLCD, computed before rounding.}
\label{tab:controls}
\centering\footnotesize
\setlength{\tabcolsep}{3pt}
\renewcommand{\arraystretch}{1.02}
\begin{tabular*}{\columnwidth}{@{\extracolsep{\fill}}lrr@{}}
\toprule
Training configuration & Avg. F1 $\uparrow$ & $\Delta$F1 \\
\midrule
\multicolumn{3}{@{}l}{\textit{(a) Training pipeline}} \\
SFT only & 60.68 & -3.65 \\
SFT $\rightarrow$ continued CE & 61.72 & -2.61 \\
SFT $\rightarrow$ adapted RLCD & 58.87 & -5.46 \\
\textbf{SFT $\rightarrow$ \method{} (full)} & \textbf{64.33} & +0.00 \\
\midrule
\multicolumn{3}{@{}l}{\textit{(b) Refinement and matched controls}} \\
Anchor only (w/o reward) & 63.70 & -0.63 \\
w/o class costs & 62.15 & -2.18 \\
w/o reference KL & 64.12 & -0.21 \\
w/ score-function estimator & 63.91 & -0.42 \\
Balanced CE only (direct) & 63.29 & -1.04 \\
Per-judgment Brier reward & 63.36 & -0.97 \\
Balanced CE (answer sequence) & 63.55 & -0.78 \\
\bottomrule
\end{tabular*}

\end{table}

\subsubsection{Component Ablations}
\label{sec:risk-ablation}

Class-balanced CE alone achieves 63.29 F1, compared with 63.70 for anchor-only training, which also retains selective reference KL. Full JA-RLCD reaches 64.33, improving over balanced CE by 1.04 points and anchor-only training by 0.63 points.

Replacing the question-group/tail reward with a per-judgment Brier reward gives 63.36 F1 while retaining the anchor, perturbations, and pathwise estimator. The 0.97-point advantage supports the combined group/tail reward formulation. Removing class costs reduces F1 to 62.15; removing reference KL or replacing pathwise differentiation with a paired score-function estimator gives 64.12 and 63.91, respectively. Appendix~\ref{app:training-controls} details these controls.

\subsubsection{Performance Across Questions}
\label{sec:question-performance}

We further examine whether JA-RLCD improves questions that are difficult for the SFT model. Using only SFT validation performance, we define the subset as the 16 lowest-F1 questions among 62 questions with sufficient validation annotations. As shown in Table~\ref{tab:question-gains}, JA-RLCD improves mean test F1 by 19.45 points over SFT on these difficult questions, compared with 6.23 points across all 82 questions. The gains remain positive relative to continued CE and anchor-only training, reaching 13.38 and 3.02 points on the difficult subset, respectively. JA-RLCD improves test F1 on 57.32\% of all questions relative to SFT. These findings show larger average gains on the validation-defined difficult subset, while improvements are not uniform across individual questions. Unlike the dataset-balanced benchmark average, these analyses weight questions equally.

\begin{table}[t]
\caption{Question-averaged test-F1 gains of JA-RLCD across all 82 questions and 16 validation-selected difficult questions.}
\label{tab:question-gains}
\centering\footnotesize
\begin{tabular*}{\columnwidth}{@{\extracolsep{\fill}}lrrr@{}}
\toprule
\method{} vs. & All Q $\Delta$F1 & Difficult Q $\Delta$F1 & Improved Q (\%) \\
\midrule
SFT & +6.23 & +19.45 & 57.32 \\
Continued CE & +4.36 & +13.38 & 56.10 \\
Anchor only & +1.38 & +3.02 & 58.54 \\
\bottomrule
\end{tabular*}

\end{table}

\subsection{Analysis}
\label{sec:analysis}
Beyond overall performance, we analyze PsyJev's ability to distinguish question-specific judgments, its inference efficiency, the reliability of its predicted probabilities, and its transfer to an unseen dataset, complemented by qualitative examples.

\subsubsection{Question-Conditioned Judgment}
\label{sec:question-conditioned}

We examine whether \model{} can distinguish different psychological judgments about the same text. Specifically, we evaluate co-annotated binary question pairs with opposite ground-truth labels, where successful prediction requires both questions to be answered correctly. We compare \model{}-4B with GLiClass, Gemma-2-9B, and the corresponding Qwen3.5-4B answer model using identical texts, questions, and semantic answer sets.

\begin{table}[t]
\caption{Question-conditioned judgment on opposite-label question pairs. Both correct and same-answer errors are reported as source-balanced percentages.}
\label{tab:question-comparison}
\centering\footnotesize
\setlength{\tabcolsep}{3pt}
\begin{tabular*}{\columnwidth}{@{\extracolsep{\fill}}lrr@{}}
\toprule
Model & Both correct $\uparrow$ & Same-answer errors $\downarrow$ \\
\midrule
GLiClass & 55.84 & 41.88 \\
Gemma-2-9B & 67.85 & 29.84 \\
Qwen3.5-4B answer & 66.24 & 31.95 \\
\textbf{\model{}-4B} & \textbf{70.06} & \textbf{28.09} \\
\bottomrule
\end{tabular*}

\end{table}

As shown in Table~\ref{tab:question-comparison}, \model{} correctly answers both questions in 70.06\% of the 35,992 pairs from eight sources, outperforming the same-size answer model (66.24\%), Gemma-2-9B (67.85\%), and GLiClass (55.84\%). Compared with the answer model, same-answer errors decrease from 31.95\% to 28.09\%, with a paired text-group 95\% confidence interval of 1.73--5.96 percentage points for the improvement in paired correctness. These results indicate stronger question-conditioned discrimination on this matched-pair subset, where the same text requires different answers under different criteria. Additional outcome breakdowns are provided in Appendix~\ref{app:question-pairs}.

\subsubsection{Inference Efficiency}
\label{sec:inference-efficiency}

\begin{figure}[t]
\centering
\captionsetup{skip=4pt}
\includegraphics[width=\columnwidth]{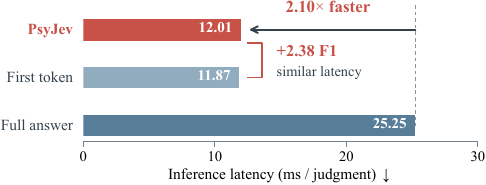}
\caption{F1--latency comparison of three 4B configurations.
PsyJev gains 2.38 F1 points over answer-SFT first-token
scoring at comparable GPU latency, while running
2.10$\times$ faster than full-answer scoring under
the shared timing protocol.}
\Description{Horizontal bars compare PsyJev at 12.01 milliseconds per judgment, first-token scoring at 11.87 milliseconds, and full-answer scoring at 25.25 milliseconds. An arrow indicates the speedup over full-answer scoring, while an annotation marks the F1 improvement over first-token scoring.}
\label{fig:scale-efficiency}
\end{figure}

Figure~\ref{fig:scale-efficiency} compares model computation time and full-test F1 for three trained 4B configurations. Full-answer and first-token scoring use the answer-SFT checkpoint, while \model{} uses its final JA-RLCD checkpoint. On the fixed 13-text training probe with CUDA Graphs on an H100 80GB, \model{} requires 12.01 ms per judgment, a $2.10\times$ speedup over full-answer scoring (25.25 ms). Its full-test F1 is 2.38 points higher than the first-token configuration at comparable model computation time. Appendix~\ref{app:timing} specifies the shared timing protocol.

\begin{table}[t]
\caption{Classification performance and probability quality on \benchmark{}. F1 and ECE are percentages. NLL is normalized by the log answer-set size. Bold indicates the best result among the displayed models.}
\label{tab:probability}
\centering\footnotesize
\setlength{\tabcolsep}{2pt}
\begin{tabular*}{\columnwidth}{@{\extracolsep{\fill}}lrrrr@{}}
\toprule
Model & F1 $\uparrow$ & NLL $\downarrow$ & Brier $\downarrow$ & ECE $\downarrow$ \\
\midrule
MentalBERT & 33.95 & 0.659 & 0.3619 & 5.82 \\
MentalRoBERTa & 38.54 & 0.620 & 0.3405 & 4.75 \\
ModernBERT-large & 52.57 & 0.580 & 0.3089 & 5.75 \\
GLiNER2 & 59.01 & 0.544 & 0.2869 & 4.95 \\
FLAN-T5-XL & 52.88 & 0.588 & 0.3220 & 6.41 \\
\midrule
\textbf{\model{}-4B} & \textbf{66.21} & \textbf{0.536} & \textbf{0.2777} & \textbf{4.58} \\
\bottomrule
\end{tabular*}

\end{table}

\begin{figure}[t]
\centering
\captionsetup{skip=4pt}
\includegraphics[width=\columnwidth]{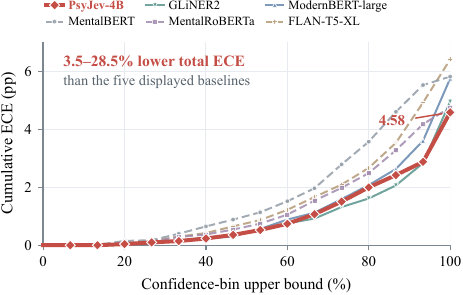}
\caption{Source-mean ECE accumulated over the full 0--100\% confidence range. Curves sum contributions from all 15 bins; their 100\% endpoints reproduce Table~\ref{tab:probability}. Lower endpoints indicate better overall calibration.}
\Description{Six curves show cumulative contributions to source-mean ECE for PsyJev-4B, GLiNER2, ModernBERT-large, FLAN-T5-XL, MentalBERT, and MentalRoBERTa. Both axes start at zero. The horizontal axis covers all 15 confidence bins from zero to 100 percent; the vertical axis measures cumulative ECE in percentage points. PsyJev-4B ends at 4.58, the lowest overall ECE among the six models, with a relative reduction of 3.5 to 28.5 percent over the five baselines.}
\label{fig:confidence}
\end{figure}

\subsubsection{Prediction Quality}
\label{sec:probability-quality}

Beyond classification accuracy, we assess the quality of predicted answer distributions using Brier score, normalized NLL, and ECE. Table~\ref{tab:probability} compares \model{}-4B with representative encoder-based, description-driven, and answer-generation models on the same test judgments, using source-balanced probability metrics (Appendix~\ref{app:confidence}). \model{}-4B achieves the highest F1 (66.21) and the lowest Brier score (0.2777) and normalized NLL (0.536) among the models displayed in Table~\ref{tab:probability}, outperforming the strongest reference, GLiNER2, on all three metrics. It also obtains the lowest ECE among these models (4.58\%) under the same source-balanced evaluation. Figure~\ref{fig:confidence} shows how source-mean calibration error accumulates across the full confidence range.

\begin{table*}[t]
\caption{Illustrative judgment corrections by \model{}-4B after JA-RLCD. $\times$ indicates disagreement with source annotations. Displayed texts are excerpts of the original test inputs.}
\label{tab:qualitative}
\centering\footnotesize
\setlength{\tabcolsep}{2pt}
\begingroup
\definecolor{caseerror}{HTML}{A64D45}

\setlength{\tabcolsep}{3pt}
\renewcommand{\arraystretch}{1.08}

\newcommand{\casecorrect}[1]{%
  \mbox{#1}%
}
\newcommand{\casewrong}[1]{%
  \textcolor{caseerror}{\mbox{#1\,\ensuremath{\times}}}%
}

\newcommand{\casefit}[2]{%
  \resizebox{#1}{!}{\mbox{#2}}%
}

\begin{tabular*}{\textwidth}{@{\extracolsep{\fill}}cccccc@{}}
\toprule

\textbf{Gold}
& GLiClass
& Gemma-2-9B
& \shortstack{Qwen3.5-4B Answer SFT}
& \shortstack{PsyJev-4B SFT}
& \cellcolor{psyjevshade}
  \shortstack{\textbf{PsyJev-4B} \textbf{JA-RLCD}} \\

\midrule

\rowcolor{black!3}
\multicolumn{6}{@{}p{\dimexpr\textwidth-2\tabcolsep\relax}@{}}{%
\strut\textbf{A. Co-occurring emotions}
\hfill
\textsc{DepressionEmo}: Multi-label emotion recognition
} \\

\multicolumn{6}{@{}p{\dimexpr\textwidth-2\tabcolsep\relax}@{}}{%
\textit{``\textbf{Very lonely} in high school, then very lonely in the military.
Eventually I was able to have a group of friends for a while and it
\textbf{filled the void}, but I ditched them,
\textbf{I get lonely now and again}''}
} \\

\addlinespace[2pt]

\textbf{\casecorrect{Emptiness, Loneliness}}
& \casewrong{Loneliness}
& \casecorrect{Emptiness, Loneliness}
& \casefit{0.19\textwidth}{\casewrong{Emptiness, Loneliness, Sadness}}
& \casefit{0.19\textwidth}{\casewrong{Emptiness, Loneliness, Sadness}}
& \cellcolor{psyjevshade}
  \textbf{\casecorrect{Emptiness, Loneliness}} \\

\addlinespace[2pt]

\multicolumn{6}{@{}p{\dimexpr\textwidth-2\tabcolsep\relax}@{}}{%
\textcolor{black!75}{%
\textbf{Interpretation.}
PsyJev recovers Emptiness and Loneliness without the additional
Sadness label predicted by the answer model and SFT.
}
} \\

\addlinespace[2pt]
\midrule

\rowcolor{black!3}
\multicolumn{6}{@{}p{\dimexpr\textwidth-2\tabcolsep\relax}@{}}{%
\strut\textbf{B. Competing distress cues}
\hfill
\textsc{CAMS}: Primary distress reason
} \\

\multicolumn{6}{@{}p{\dimexpr\textwidth-2\tabcolsep\relax}@{}}{%
\textit{``i have a \textbf{part-time job} and
\textbf{struggle to even get through 4 hour shifts}.
so i \textbf{don't make enough money} to do anything
but attend work basically and smoke some cigarettes.
[...] i'm also struggling with a
\textbf{severely broken heart} from an on-and-off 10 year thing''}
} \\

\addlinespace[2pt]

\textbf{\casecorrect{Jobs and careers}}
& \casewrong{Relationship}
& \casewrong{Relationship}
& \casewrong{Relationship}
& \casewrong{Relationship}
& \cellcolor{psyjevshade}
  \textbf{\casecorrect{Jobs and careers}} \\

\addlinespace[2pt]

\multicolumn{6}{@{}p{\dimexpr\textwidth-2\tabcolsep\relax}@{}}{%
\textcolor{black!75}{%
\textbf{Interpretation.}
Limited work capacity and insufficient income are consistent with the annotated Jobs and careers category, despite the competing relationship cue.
}
} \\

\addlinespace[2pt]
\midrule

\rowcolor{black!3}
\multicolumn{6}{@{}p{\dimexpr\textwidth-2\tabcolsep\relax}@{}}{%
\strut\textbf{C. Related cognitive patterns}
\hfill
\textsc{TherapistQA}: Cognitive distortion category
} \\

\multicolumn{6}{@{}p{\dimexpr\textwidth-2\tabcolsep\relax}@{}}{%
\textit{``My ss is 17. I've been in his life since age 6.
He has \textbf{always lied and been spiteful}.
[...] \textbf{The lying has never stopped}.
He just got better at it.
He learned how to lie with sincerity.''}
} \\

\addlinespace[2pt]

\textbf{\casecorrect{Overgeneralization}}
& \casewrong{No Distortion}
& \casecorrect{Overgeneralization}
& \casecorrect{Overgeneralization}
& \casewrong{Labeling}
& \cellcolor{psyjevshade}
  \textbf{\casecorrect{Overgeneralization}} \\

\addlinespace[2pt]

\multicolumn{6}{@{}p{\dimexpr\textwidth-2\tabcolsep\relax}@{}}{%
\textcolor{black!75}{%
\textbf{Interpretation.}
The unqualified always/never statements support the annotated
Overgeneralization category; Labeling is a related competing
interpretation selected by SFT.
}
} \\

\addlinespace[2pt]
\bottomrule

\end{tabular*}
\endgroup
\end{table*}
\subsubsection{Unseen-Dataset Transfer}
\label{sec:unseen-transfer}

We evaluate transfer to Dreaddit, an unseen stress-detection dataset with 715 test examples excluded from \benchmark{} supervision \citep{turcan2019dreaddit}. As shown in Table~\ref{tab:dreaddit}, \model{} achieves 77.76 F1 at 2B and 78.70 at 4B, improving over its direct-judgment SFT initialization by 3.85 and 0.83 points, respectively. Both variants also outperform their corresponding answer-SFT baselines by 3.44 and 2.17 points. The 4B result is competitive with Gemma-3-4B (78.26) and Llama-3.2-3B (78.00). These results suggest that the benefits of judgment-anchored refinement can extend beyond the benchmark's training datasets, with gains varying by model scale. Appendix~\ref{app:dreaddit} reports the complete comparison.

\begin{table}[t]
\caption{Dreaddit transfer on all 715 test examples (F1, \%). No Dreaddit supervision is used.}
\label{tab:dreaddit}
\centering\footnotesize
\setlength{\tabcolsep}{3pt}
\begin{tabular*}{\columnwidth}{@{\extracolsep{\fill}}llr@{}}
\toprule
Scale & Model / training & F1 (\%) \\
\midrule
2B & Qwen3.5 answer-SFT & 74.32 \\
   & \model{} SFT & 73.91 \\
   & \model{} JA-RLCD & \textbf{77.76} \\
\midrule
4B & Qwen3.5 answer-SFT & 76.53 \\
   & \model{} SFT & 77.87 \\
   & \model{} JA-RLCD & \textbf{78.70} \\
\midrule
3B & Llama-3.2-3B & 78.00 \\
4B & Gemma-3-4B & 78.26 \\
\bottomrule
\end{tabular*}

\end{table}

\subsubsection{Qualitative Analysis}
\label{sec:qualitative}

Table~\ref{tab:qualitative} presents three test cases in which JA-RLCD corrects predictions from its SFT initialization. The cases cover co-occurring emotions, competing sources of distress, and closely related cognitive distortion categories. In Case A, \model{}-4B recovers the annotated emotion set without the additional Sadness label predicted by SFT. In Case B, it identifies employment-related distress despite concurrent relationship difficulties. In Case C, it selects the annotated Overgeneralization category, whereas SFT favors Labeling. The answer model also selects the gold category. These examples illustrate how JA-RLCD can improve label selection across different psychological judgment criteria. Appendix~\ref{app:cases} shows the corresponding SFT-to-JA-RLCD probability shifts.

\paragraph{Scope of use.}
These judgments describe annotated textual content and do not establish clinical diagnoses. Reuse should follow each source dataset's access and usage terms and safeguard the privacy of sensitive self-disclosures.

\section{Conclusion}
We presented \benchmark{} and \model{} for unified psychological
analysis of Web content through question-conditioned judgments.
\benchmark{} organizes heterogeneous tasks while retaining their
distinct decision criteria, and \model{} enables direct probabilistic
prediction with \method{} for judgment-anchored refinement.
Across 2B, 4B, and 9B scales, \model{} achieves average F1 scores
of 64.33, 66.21, and 67.56, respectively, outperforming the
corresponding answer-SFT baselines.
Additional analyses examine question-conditioned discrimination,
inference efficiency, and probability quality.
These results demonstrate the value of a shared, question-conditioned
approach to heterogeneous psychological content analysis.
\FloatBarrier
\label{last-main-page}

\bibliographystyle{ACM-Reference-Format}
\bibliography{references}

\appendix

\section{Data Construction and Splits}
\label{app:data}

\begin{table}[!ht]
\caption{Source records and coordinated splits. Q counts registry questions: 79 binary and three multiclass in total.}
\label{tab:data}
\centering\footnotesize
\setlength{\tabcolsep}{3pt}
\renewcommand{\arraystretch}{1.02}
\begin{tabular*}{\columnwidth}{@{\extracolsep{\fill}}lrrrr@{}}
\toprule
Source & Q & Train & Val. & Test \\
\midrule
\multicolumn{5}{@{}l}{\textit{Depression / anxiety}}\\
DepressionEmo & 8 & 4,219 & 907 & 911 \\
ANGST & 2 & 6,769 & 763 & 2,872 \\
PRIMATE & 9 & 1,593 & 207 & 203 \\
BDI-Sen & 21 & 1,759 & 261 & 503 \\
\addlinespace[2pt]
\multicolumn{5}{@{}l}{\textit{Self / relationships}}\\
IRF & 2 & 1,267 & 483 & 1,771 \\
LoST & 1 & 740 & 271 & 1,163 \\
LonXplain & 1 & 1,267 & 483 & 1,771 \\
\addlinespace[2pt]
\multicolumn{5}{@{}l}{\textit{Stress / distress}}\\
SAD & 1 & 5,533 & 655 & 662 \\
CAMS & 1 & 2,472 & 643 & 1,920 \\
\addlinespace[2pt]
\multicolumn{5}{@{}l}{\textit{Cognition}}\\
Thinking Traps & 15 & 537 & 68 & 93 \\
TherapistQA & 1 & 2,023 & 245 & 260 \\
\addlinespace[2pt]
\multicolumn{5}{@{}l}{\textit{Wellness}}\\
MultiWD & 6 & 1,155 & 444 & 1,682 \\
\addlinespace[2pt]
\multicolumn{5}{@{}l}{\textit{Crisis}}\\
CRADLE & 14 & 4,177 & 422 & 601 \\
\midrule
Total & 82 & 33,511 & 5,852 & 14,412 \\
\bottomrule
\end{tabular*}

\end{table}

\paragraph{Label construction.}
Missing labels are masked. BDI-Sen retains symptom relevance, excluding severity. TherapistQA uses the patient question and dominant distortion, excluding therapist responses, spans, and secondary labels. Thinking Traps merges normalized repeated completions into 15-category label sets; CRADLE expands seven crisis categories across ongoing/past status into 14 questions. Missing CAMS categories remain unobserved. Table~\ref{tab:registry-examples} gives four verbatim registry entries.

\begin{table}[t]
\caption{Verbatim registry instructions and legal answers for four judgment types in \benchmark{}.}
\label{tab:registry-examples}
\centering\fontsize{7.5}{8.8}\selectfont
\setlength{\tabcolsep}{2pt}
\renewcommand{\arraystretch}{1.1}
\begin{tabular}{@{}>{\raggedright\arraybackslash}p{.21\columnwidth} >{\raggedright\arraybackslash}p{.75\columnwidth}@{}}
\toprule
Source & Registry instruction and legal answers \\
\midrule
DepressionEmo & Does the author express sadness in this post?\par\emph{Answers:} no; yes \\
\addlinespace[4pt]
PRIMATE & Does this post provide enough information to answer the PHQ-9 item about Feeling down depressed or hopeless? Judge answerability, not whether the symptom is present.\par\emph{Answers:} no; yes \\
\addlinespace[4pt]
BDI-Sen & Does this sentence provide information relevant to the BDI-II symptom Sadness? Judge the source relevance label, not severity or symptom presence.\par\emph{Answers:} no; yes \\
\addlinespace[4pt]
SAD & What is the primary stressor category expressed in this sentence?\par\emph{Answers:} Financial Problem; Other; Everyday Decision Making; Emotional Turmoil; School; Family Issues; Social Relationships; Work; Health, Fatigue, or Physical Pain \\
\bottomrule
\end{tabular}

\end{table}

\paragraph{Split coordination.}
Exact texts and available source groups are coordinated across all 13 sources, with test taking priority over validation and training. Normalization applies Unicode NFKC, case folding, and whitespace collapse. Shared texts retain their distinct source-defined judgments. We mask 22 conflicting judgments across 11 identical text--target keys.

\paragraph{TherapistQA refinement.}
After removing two identical questions with conflicting labels, normalized-question groups receive a seed-42 hash split, inheriting exact-match assignments. Candidates use up to 12 smallest distinct CRC32 four-token shingle hashes and the first 128 indexed texts per hash. Components satisfying the Jaccard or containment criteria in Table~\ref{tab:split-overlap} follow the same held-out priority; the other 12 sources remain unchanged. ANGST retains silver training/validation and expert-labeled test annotations; CRADLE's coordinated test partition contains 600 gold and one silver record.

\paragraph{Final-split audit.}
The frozen snapshot contains 53,775 records and 41,349 normalized texts. All 3,600 cross-source shared-text groups are confined to one split (training: 1,301; validation: 494; test: 1,805), including all 3,518 IRF--LonXplain shared texts. Table~\ref{tab:split-overlap} checks every partition pair using uncapped prefix filtering and exact four-word shingles, on both full texts and their first 512 tokens. No source-group identifier crosses partitions. The audit measures lexical instance reuse.

\begin{table}[t]
\caption{Cross-partition overlap on full texts and 512-token inputs.}
\label{tab:split-overlap}
\centering\footnotesize
\setlength{\tabcolsep}{3pt}
\begin{tabular*}{\columnwidth}{@{\extracolsep{\fill}}lrr@{}}
\toprule
Cross-partition check & Full text & 512 tokens \\
\midrule
Exact normalized matches & 0 (0.00\%) & 0 (0.00\%) \\
Jaccard $\geq0.85$, $\geq8$ shingles & 0 & 0 \\
Containment ($\geq80$ characters) & 0 & 0 \\
\bottomrule
\end{tabular*}

\end{table}

\section{Reproducibility and Comparison Protocols}
\label{app:implementation}
\subsection{Input Processing}
We use \texttt{Qwen/Qwen3.5-2B}, \texttt{Qwen/Qwen3.5-4B}, and
\texttt{Qwen/Qwen3.5-9B} (revision prefixes \texttt{15852e8c},
\texttt{851bf6e8}, and \texttt{c2022362}), each with its native
tokenizer. Input texts are encoded without added special tokens
and truncated to 512 tokens, while question criteria and legal
answer descriptions are preserved in full and in registry order.
Requests serialize the text under \texttt{Context:}, followed by
\texttt{Question:} and letter-prefixed answer options. We apply
the backbone chat template with thinking disabled and append
a separately tokenized \texttt{Answer:} prefix for constrained
readout. For multilabel sources, each observed binary question
is scored separately; a positive label is assigned only when
$p(\mathrm{yes}) > p(\mathrm{no})$, with ties assigned to no.

\subsection{Training and Matched Controls}
\paragraph{Optimizer and accumulation.}
All reported runs use seed 42. Each stage visits all 33,511 training records once in 524 optimizer updates, with effective batches of 63 or 64 records, retaining its final checkpoint. All observed questions are retained in record/question order, grouped by source, and split into consecutive source-homogeneous microbatches of at most 16 judgments. Their weighted gradients are summed over each complete record batch before one optimizer step; the microbatch count varies with the observed labels. AdamW uses $\beta_1=0.9$, $\beta_2=0.999$, $\epsilon=10^{-8}$, weight decay 0.01, global gradient clipping at 1, and 3\% warmup followed by cosine decay. The 2B model uses full tuning at $2\times10^{-6}$ in both stages. At 4B/9B, LoRA uses rank 16, scaling 32, and zero dropout, targeting attention projections (\texttt{q/k/v/o\_proj}), linear-attention input/output projections (\texttt{in\_proj\_qkv}, \texttt{in\_proj\_z}, \texttt{in\_proj\_b}, \texttt{in\_proj\_a}, and \texttt{out\_proj}), and MLP gate, up, and down projections. The vocabulary head remains frozen; all trainable adapters share learning rates of $2\times10^{-4}$ for SFT and $2\times10^{-6}$ for refinement.

\paragraph{Weights and optimization.}
For training counts $n_{qk}$ and $n_q=\sum_k n_{qk}$, we set
\begin{equation}
\widehat\pi_{qk}=\frac{n_{qk}+1}{n_q+K_q},\qquad
c_{qk}=\min(8,\eta_q/\widehat\pi_{qk}),
\end{equation}
where $\eta_q$ ensures $\sum_k\widehat\pi_{qk}c_{qk}=1$. For $N$ training records, $S$ sources, effective record batch size $B$, and source question set $\mathcal Q_s$,
\begin{equation}
a_i=\frac{N}{B S |\mathcal Q_{s_i}|n_{q_i}}.
\end{equation}

The anchor applies $a_i$ per judgment to balance source/question observation frequency; the reward branch uses only their sum $A_{\mathcal B}$ in Eq.~\ref{eq:smoothing}. Priors and weights derive from training counts;
reference-correctness gates also use training labels only. Refinement initializes from the same-scale final SFT checkpoint, keeping a frozen reference copy. Each microbatch uses two Gaussian directions and their negatives, with all four reward evaluations sharing the same model logits. The noise scale decreases linearly from $0.4$ to $0.1$ over refinement updates; $\tau=4$ and $\lambda=0.05$. With $u$ the pathwise reward gradient, the update accumulates $g-u$ over microbatches and clips the total before AdamW. The refinement configuration was selected using validation only, without test results.

\subsection{Baseline Protocols}
\label{app:baselines}

All benchmark-adapted baselines use the same cleaned records, question registry, and observed-label masks. Fixed-head classifiers use question-specific output heads, while description-driven models preserve the original decision criteria in their input schemas. Answer-SFT models score complete legal-answer sequences, including EOS. All models use their final training checkpoints, except ModernBERT, which selects among three epochs based on validation F1, followed by NLL and earlier-epoch tie-breaking. MentaLLaMA undergoes benchmark-specific answer SFT from its existing mental-health instruction-tuned checkpoint \citep{yang2024mentallama}.

\paragraph{Open-source Laya.}
We reproduce Laya's SFT and RLCD training pipeline on \benchmark{} using the released \href{https://huggingface.co/convaiinnovations/laya-multilingual/tree/1720e3e3357cfe1e281542e223f8273b0890ca34}{Laya-multilingual checkpoint} (revision \texttt{1720e3e3}), preserving its native architecture and RLCD objective. We first perform one epoch of full-parameter SFT (524 updates, learning rate $10^{-5}$), followed by one epoch of RLCD (524 updates, learning rate $2\times10^{-6}$). RLCD combines cross-entropy with score-function optimization over Gaussian-perturbed answer logits, using four samples, a noise schedule from $0.4$ to $0.1$, and spherical/ranked-probability reward weights of $0.75/1.0$. Both stages use the benchmark training partition and are evaluated using their final checkpoints.

\paragraph{Jev API evaluation.}
We evaluate \texttt{jev-1.14.0} through Typesafe's official API on all 44,999 test judgments. Each request includes the source text, question criterion, and legal answers, with the source text capped at 512 Qwen tokens. Binary and multiclass questions use Noul and Choice, respectively. The API is evaluated zero-shot without benchmark fine-tuning or demonstrations, and its returned probabilities are aligned with the benchmark's semantic labels. All 84 invalid responses remain in the evaluation as incorrect predictions; the invalid label is excluded from the legal-class macro average when computing F1. Results follow the same question- and source-level aggregation as the main evaluation.

\paragraph{Matched refinement controls.}
\label{app:training-controls}
\label{app:question-gains}
All direct-scoring 2B continuations start independently from the same final SFT checkpoint and retain the same records, registry, observed-label masks, record schedule, trainable parameters, optimizer, and 524 additional updates. Adapted RLCD combines ordinary CE with Laya's categorical score-function objective \citep{laya2026}: four independent centered Gaussian logit actions, advantages centered and normalized by their standard deviation, and reward weights of 0.75 for the spherical term and 1.0 for ranked probability. It uses SFT source/record weights without the JA reference gate or question-group aggregation.

\paragraph{Ablation configurations.}
The anchor-only variant removes reward optimization while retaining balanced supervision and selective reference KL. Removing class costs sets $c_{qk}=1$ in both the anchor and reward objectives, while removing reference KL sets $\lambda=0$. The score-function variant replaces pathwise differentiation with paired score-function gradients, retaining the same anchor, question-group reward, and logit-space perturbations.

The balanced-CE-only control disables both reference KL and reward optimization, retaining the source/question weights and class costs. The ordinary Brier control replaces the group/tail reward with one minus the class-cost-normalized mean per-judgment half-Brier error, retaining the anchor, perturbations, reward scale, and pathwise estimator. This changes grouping and tail weighting jointly. The answer-sequence balanced-CE control starts from its corresponding answer-SFT checkpoint and uses full-vocabulary CE on the answer letter and EOS. It matches the direct balanced-CE control's pretrained backbone, record schedule, weights, optimizer, and two total training epochs, including 524 continuation updates.

\subsection{Unified Timing Protocol}
\label{app:timing}
\begin{figure}[t]
\centering
\captionsetup{skip=4pt}
\includegraphics[width=\columnwidth]{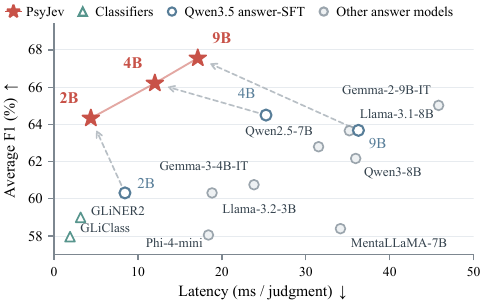}
\caption{Quality--latency comparison of the 16 models
with average F1 $\geq 57\%$ in Table~\ref{tab:main}.}
\Description{A scatter plot focuses on models with average F1 of at least 57 percent and uses a linear latency axis. It shows PsyJev at 2B, 4B, and 9B, alongside thirteen baselines. Dashed arrows pair PsyJev and Qwen3.5 answer-SFT at matching backbone sizes.}
\label{fig:appendix-scale-efficiency}
\end{figure}

Model computation time is measured on a shared probe comprising 13
training texts (the second record in each source's stored training
order) and 64 observed judgments. Each request contains one text
and its observed questions and is repeated five times after warmup.
All models run on a single NVIDIA H100 80GB using BF16 and CUDA
Graphs. CUDA events measure GPU execution, including text prefill
and answer scoring, while excluding loading, tokenization, data
transfer, and graph capture. Inputs reside on the GPU, and latency
is reported as total measured time divided by the number of
judgments. Each model retains its native scoring mechanism under
the same protocol. For the matched 4B comparison, full-answer and first-token scoring share the Qwen3.5-4B answer-SFT checkpoint, while \model{} uses its
final JA-RLCD checkpoint. Their respective latencies are 25.25,
11.87, and 12.01 ms per judgment. Figure~\ref{fig:scale-efficiency}
pairs these measurements with full-test F1, while
Figure~\ref{fig:appendix-scale-efficiency} compares quality and
latency across the models in Table~\ref{tab:main}.

\section{Prediction Diagnostics}
\label{app:diagnostics}

\subsection{Probability Evaluation Protocol}
\label{app:confidence}

All probability evaluations use the same 44,999 test judgments and legal answer sets. Classification models normalize their answer scores over the legal options, while answer-generation models normalize complete answer-sequence likelihoods, including EOS. No post-hoc calibration is applied. For a judgment with $K$ answers, normalized NLL and Brier score are defined as:
\begin{equation}
\mathrm{NLL}=-\frac{\log p_y}{\log K},
\qquad
\mathrm{Brier}=\sum_{k=1}^{K}(p_k-\mathbb I[k=y])^2.
\end{equation}

NLL is averaged over judgments within records and then equally across records and sources. Brier score and 15-bin ECE are computed at the source level and averaged equally across sources.

For Figure~\ref{fig:confidence}, let $n_{s,b}$, $a_{s,b}$, and $c_{s,b}$ denote the count, accuracy, and mean confidence in source $s$ and bin $b$. The source-bin contribution and cumulative curve are
\[
e_{s,b}=\frac{n_{s,b}}{n_s}|a_{s,b}-c_{s,b}|,
\qquad
E_{\le b}=\frac{1}{13}\sum_s\sum_{j=1}^{b}e_{s,j}.
\]
Empty bins contribute zero. The denominator $n_s$ is the full source judgment count, so $E_{\le15}$ equals the source-mean ECE in Table~\ref{tab:probability}.

\begin{table}[t]
\caption{Probability quality of the final PsyJev-4B by source. The mean gives equal weight to all 13 sources.}
\label{tab:probability-sources-4b}
\centering\footnotesize
\setlength{\tabcolsep}{6pt}
\begin{tabular}{@{}lrr@{}}
\toprule
Source & Brier $\downarrow$ & ECE (\%) $\downarrow$ \\
\midrule
ANGST & 0.4436 & 18.54 \\
BDI-Sen & 0.1499 & 1.85 \\
CAMS & 0.5408 & 4.33 \\
CRADLE & 0.0644 & 0.76 \\
DepressionEmo & 0.2327 & 1.06 \\
IRF & 0.1895 & 3.97 \\
LoST & 0.1889 & 3.41 \\
LonXplain & 0.2179 & 6.06 \\
MultiWD & 0.2143 & 0.84 \\
PRIMATE & 0.2444 & 2.35 \\
SAD & 0.2997 & 4.34 \\
TherapistQA & 0.6560 & 7.18 \\
Thinking Traps & 0.1676 & 4.86 \\
\midrule
Source mean & 0.2777 & 4.58 \\
\bottomrule
\end{tabular}

\end{table}

\subsection{Unseen-Dataset Transfer}
\label{app:dreaddit}
All models use their benchmark-trained checkpoints on the complete 715-example Dreaddit test set~\citep{turcan2019dreaddit}, without Dreaddit supervision or further tuning. The registered question is \emph{Does the author express psychological stress in this text?} Legal answers are ordered (\texttt{no}, \texttt{yes}), mapping original labels 0 and 1, respectively. Texts retain their first 512 tokens; prediction uses legal-answer argmax, and F1 treats \texttt{yes} as positive. Auditing all 715 texts against the actual \benchmark{} training (33,511) and validation (5,852) records found no raw/normalized exact matches, no word-level 4-gram Jaccard matches $\geq0.85$ with uncapped candidate retrieval, and no full-text containments with the shorter text at least 80 characters. Table~\ref{tab:dreaddit-full} lists all 12 configurations.

\begin{table}[t]
\caption{Complete Dreaddit transfer results (F1, \%) without additional training on Dreaddit.}
\label{tab:dreaddit-full}
\centering\footnotesize
\setlength{\tabcolsep}{3pt}
\begin{tabular*}{\columnwidth}{@{\extracolsep{\fill}}lr@{}}
\toprule
Model / training & F1 (\%) \\
\midrule
GLiClass & 66.56 \\
GLiNER2 & 70.94 \\
FLAN-T5-XL & 74.84 \\
Llama-3.2-3B & 78.00 \\
Phi-4-mini & 73.27 \\
Gemma-3-4B & 78.26 \\
\midrule
Qwen3.5-2B answer-SFT & 74.32 \\
\model{}-2B SFT & 73.91 \\
\model{}-2B JA-RLCD & \textbf{77.76} \\
\midrule
Qwen3.5-4B answer-SFT & 76.53 \\
\model{}-4B SFT & 77.87 \\
\model{}-4B JA-RLCD & \textbf{78.70} \\
\bottomrule
\end{tabular*}

\end{table}

\begin{table}[!ht]
\caption{Different questions about the same text. Outcomes (\%) use identical opposite-label binary pairs and equal source weights. ``Same answer'' means exactly one prediction is wrong in this subset.}
\label{tab:paired-behavior}
\centering\footnotesize
\setlength{\tabcolsep}{2pt}
\begin{tabular*}{\columnwidth}{@{\extracolsep{\fill}}lrrr@{}}
\toprule
Model & Both correct $\uparrow$ & Same answer & Both wrong \\
\midrule
GLiClass & 55.84 & 41.88 & 2.28 \\
Gemma-2-9B & 67.85 & 29.84 & 2.31 \\
Qwen3.5-4B answer & 66.24 & 31.95 & 1.82 \\
\model{}-4B SFT & 63.55 & 34.37 & 2.09 \\
\textbf{\model{}-4B} & 70.06 & 28.09 & 1.84 \\
\bottomrule
\end{tabular*}

\end{table}

\subsection{Questions about Shared Text}
\label{app:question-pairs}
Table~\ref{tab:paired-behavior} adds \model{}-4B SFT and both-wrong
outcomes to the main comparison. All models use identical
opposite-label binary pairs, with three outcomes.
Results are source-balanced, with paired text-group resampling
accounting for shared-text dependence.

\subsection{Case-Level Probability Shifts}
\label{app:cases}
Table~\ref{tab:case-probabilities} follows the three 4B cases in Table~\ref{tab:qualitative}, using the final JA-RLCD checkpoint and its SFT parent. In Case A, the Sadness probability crosses below 0.5 while Emptiness and Loneliness remain positive. In Cases B and C, the gold label overtakes the strongest competing label; both labels' probabilities are shown to distinguish relative preference changes from uniform confidence increases. These cases use DepressionEmo:16ll3jf, CAMS:1286, and TherapistQA:364. Models receive the original inputs; display excerpts preserve their order, mark omissions, and normalize punctuation.
\begin{table}[ht]
\caption{Answer probabilities before and after JA-RLCD for the three cases in Table~\ref{tab:qualitative}.}
\label{tab:case-probabilities}
\centering\footnotesize
\setlength{\tabcolsep}{3pt}
\begin{tabular*}{\columnwidth}{@{\extracolsep{\fill}}llrr@{}}
\toprule
Case & Answer probability / margin & SFT & JA-RLCD \\
\midrule
A & Emptiness & 0.752 & 0.825 \\
A & Loneliness & 0.999 & 1.000 \\
A & Sadness & 0.571 & 0.445 \\
\addlinespace[3pt]
B & Jobs and careers & 0.303 & 0.402 \\
B & Relationship & 0.406 & 0.389 \\
B & $p_y-p_c$ & -0.103 & +0.013 \\
\addlinespace[3pt]
C & Overgeneralization & 0.319 & 0.433 \\
C & Labeling & 0.336 & 0.414 \\
C & $p_y-p_c$ & -0.017 & +0.019 \\
\bottomrule
\end{tabular*}

\end{table}

\end{document}